\documentclass{article} 
\usepackage{iclr2026_conference,times}

\usepackage{amsmath,amsfonts,bm}

\def\eqref#1{equation~\ref{#1}}

\def\1{\bm{1}}

\DeclareMathAlphabet{\mathsfit}{\encodingdefault}{\sfdefault}{m}{sl}
\SetMathAlphabet{\mathsfit}{bold}{\encodingdefault}{\sfdefault}{bx}{n}

\usepackage{url}
\usepackage{listings}
\usepackage[dvipsnames]{xcolor}
\usepackage{amssymb} 
\usepackage{graphicx}
\usepackage{pifont}
\usepackage{booktabs}
\usepackage{multirow}
\usepackage[textsize=tiny]{todonotes}
\usepackage{subcaption}

\usepackage[
  colorlinks=true,
  citecolor=NavyBlue,
  linkcolor=NavyBlue,
  urlcolor=NavyBlue
]{hyperref}
\usepackage{cleveref}

\newcommand{\cmark}{\textcolor{green!60!black}{\ding{51}}}
\newcommand{\xmark}{\textcolor{red!70!black}{\ding{55}}} 

\title{{\rm stable-worldmodel}-v1: Reproducible World Modeling Research and Evaluation}

\author{%
Lucas Maes\textnormal{*\textsuperscript{1}}~~ Quentin Le Lidec\textnormal{*\textsuperscript{2}}~~
Dan Haramati\textnormal{\textsuperscript{3}}~~
Nassim Massaudi\textnormal{\textsuperscript{4}}
\\[3pt]
\textbf{Damien Scieur\textnormal{\textsuperscript{1,5}}~~ Yann LeCun\textnormal{\textsuperscript{2}}~~Randall Balestriero\textnormal{\textsuperscript{3}}}
\\[8pt]
$^{1}$Mila \& Université de Montréal~~$^{2}$New York University\\$^{3}$Brown University~~$^{4}$Independent Researcher~~$^{5}$Samsung SAIL
}
\iclrfinalcopy 
\begin{document}

\renewcommand{\thefootnote}{}
\footnotetext{* Equal contribution. Correspondence to \texttt{lucas.maes@mila.quebec}}
\renewcommand{\thefootnote}{\arabic{footnote}}

\maketitle

\begin{abstract}
World Models have emerged as a powerful paradigm for learning compact, predictive representations of environment dynamics, enabling agents to reason, plan, and generalize beyond direct experience.
Despite recent interest in World Models, most available implementations remain publication-specific, severely limiting their reusability, increasing the risk of bugs, and reducing evaluation standardization. To mitigate these issues, we introduce \texttt{stable-worldmodel} ({\rm SWM}), a modular, tested, and documented world-model research ecosystem that provides efficient data-collection tools, standardized environments, planning algorithms, and baseline implementations. 
In addition, each environment in {\rm SWM} enables controllable factors of variation, including visual and physical properties, to support robustness and continual learning research. 
Finally, we demonstrate the utility of {\rm SWM} by using it to study zero-shot robustness in DINO-WM.
\end{abstract}

\begin{quote}
\centering
\textcolor{Gray}{-- \emph{World Model Research Made Simple.}}
\end{quote}

\section{Introduction}
\label{sec:introduction}

A promising paradigm toward building capable and general-purpose embodied agents involves learning dynamics models of the world, commonly referred to as World Models (WM,~\citet{ha2018world}). Despite rapid progress and growing community interest, research on WMs remains fragmented and lacks shared benchmarks comparable to those in vision~\citep{russakovsky2015imagenet,lin2014microsoft}, reinforcement learning~\citep{bellemare2013arcade, gym, tassa2018deepmind}, or language modeling~\citep{wang2024mmlu, phan2025humanity}. This diversity of paradigms, design choices, and environments complicates meaningful comparison between methods. Systematic re-implementation of utilities further exacerbates this issue: for example, two recent works, PLDM \citep{sobal2025stresstesting} and DINO-WM\citep{zhou2025dino-wm}, re-implement the same Two-Room environment with substantial divergence (81 deletions, 86 additions, and 18 updates), underscoring the lack of shared infrastructure. Moreover, beyond comparing performance across disparate environments, controlled variations within a single environment are essential to isolate key factors, probe generalization, and better understand the inductive biases and failure modes of WMs.

In this work, we introduce \texttt{stable-worldmodel}, a new research ecosystem designed to facilitate streamlined and reproducible experimentation and benchmarking WMs. We design a simple, easy-to-use API that allows custom dataset collection, training, and evaluation, as well as integration of novel algorithms and environments to support future growth and development. A comparison with other recent latent world model codebases is provided in Table \ref{tab:wm_library_comparison}.

\begin{table}[h]
\centering
\small
\caption{\small\textbf{Latent World-Model codebases comparison.} (PR = Pull Request, LoC = Lines of Code) Collected statistics demonstrate the lack of a reliable, open-source, and unified codebase to perform world model research. We address this issue with our proposed library {\rm SWM}.}
\label{tab:wm_library_comparison}
\centering
\small
\setlength{\tabcolsep}{4pt}
\begin{tabular}{@{}lccc@{}}
\toprule
 & \textbf{{\rm SWM}} (ours) & \textbf{PLDM} & \textbf{DINO-WM} \\
\midrule
Backend        & PyTorch & PyTorch & PyTorch \\
Documentation  & \cmark  & \xmark  & \xmark \\
\# Baselines   & 4       & 1       & 1 \\
\# Environments& 16      & 2       & 4 \\
\# FoV (per env)& 6-17     & 0       & 0 \\
Type Checking  & \cmark  & \cmark  & \xmark \\
Test Coverage  & 73\%    & 0\%     & 0\% \\
Last Commit    & $<$1 week  & $>$3 months & $>$10 months \\
PRs (6 mo.)    & 99      & 1       & 0 \\
\# LoC         & 3562    & 6796    & 4349 \\
\bottomrule
\end{tabular}
\end{table}

\section{Stable World Model Ecosystem: An Overview}
\label{sec:overview}

Stable World Model ({\rm SWM}) goal is to support researchers by reducing the idea-to-experiment time gap. We build the library around the philosophy that people already have their codebase or tool for training their model. Therefore, our library should focus on providing support for their training with a ready-to-use environment and utilities for data collection or model evaluation. In the rest of this section, we provide an overview of the user API and the different components of the library. A full overview of a typical world model pipeline with {\rm SWM} is provided in Listing \ref{lst:swm-pusht}.

\subsection{The World interface: streamlined WM research}

\begin{lstlisting}[caption={\small\textbf{World Interface Logic.} After specifying the environment ID (e.g., \texttt{swm/PushT-v1}) and the number of simulations, a policy can be attached to enable online interaction with the environment. At any time, all simulation-related information can be accessed via the \texttt{infos} dictionary.}, label={lst:swm-world}]
    import stable_worldmodel as swm
    
    world = swm.World('swm/PushT-v1', num_envs=8)
    world.set_policy(YourExpertPolicy())
    
    world.reset() # initialize the world
    world.step()  # update the world state with policy
    world.infos   # current world state (dict)
\end{lstlisting}

The core abstraction in {\rm SWM} is the \texttt{World}. A \texttt{World} wraps one or more Gymnasium \cite{tow} environments and provides a unified interface for simulation, data collection, debugging, and evaluation. Internally, it leverages Gymnasium’s synchronous environment API to manage and step multiple environments within a single object.

Unlike the widely used Gymnasium \citep{towers2025gymnasium} interface, a \texttt{World} does not return observations, rewards, or termination flags from \texttt{reset} or \texttt{step}. Instead, all data produced by the environments is stored in a single internal dictionary, \texttt{world.infos}, which is updated in place at every reset or step. Both methods operate synchronously over all environments, making the complete simulation state accessible at any time via \texttt{world.infos}.

Action selection in {\rm SWM} is handled by a policy object attached to the \texttt{World}. The \texttt{step} method does not take actions as input; instead, at each step, the world queries its policy to obtain actions for all environments. A policy is a lightweight Python object implementing a \texttt{get\_action} method, which takes the current \texttt{world.infos} as input and returns one action per environment. This design cleanly decouples control logic from environment execution, allowing policies to be swapped without modifying the world interface. 

Once a policy is attached to a \texttt{World}, it can be used to record datasets or perform evaluation. Dataset recording executes the policy over episodes and logs all information contained in \texttt{world.infos}, while evaluation runs the same execution loop without data persistence. In both cases, the behavior and properties of the resulting trajectories are entirely determined by the chosen policy and world configuration. An illustrative example of dataset recording is provided in Listing \ref{lst:swm-fov}. Additional details about the dataset and evaluation are reported in Appendix \ref{appendix:swm-details}.

\subsection{Environments and Factor of Variations}

\begin{figure}
    \centering
    \begin{minipage}[t]{0.19\textwidth}
      \centering
     \includegraphics[width=\linewidth]{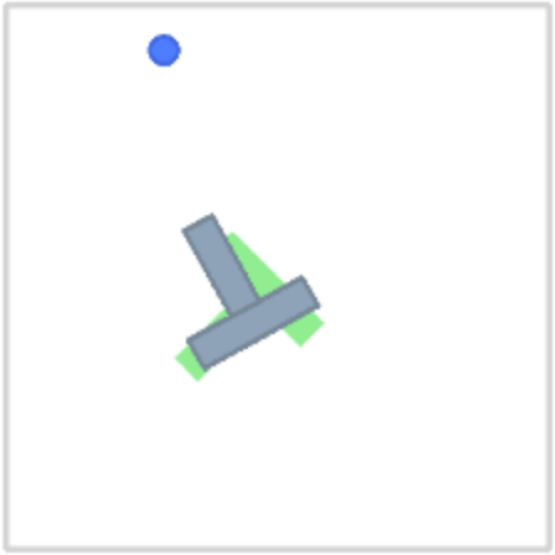}
     \includegraphics[width=\linewidth]{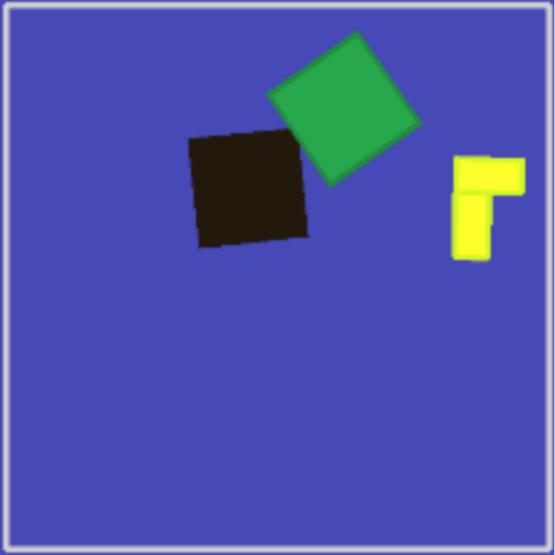}
      (a) PushT
    \end{minipage}\hspace{0.3em}
    \begin{minipage}[t]{0.19\textwidth}
      \centering
     \includegraphics[width=\linewidth]{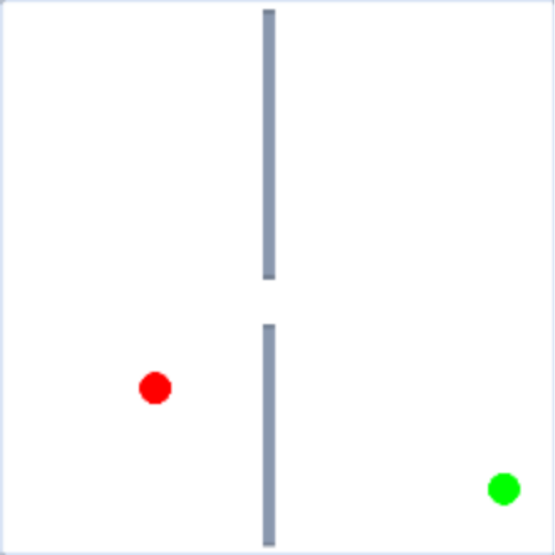}
     \includegraphics[width=\linewidth]{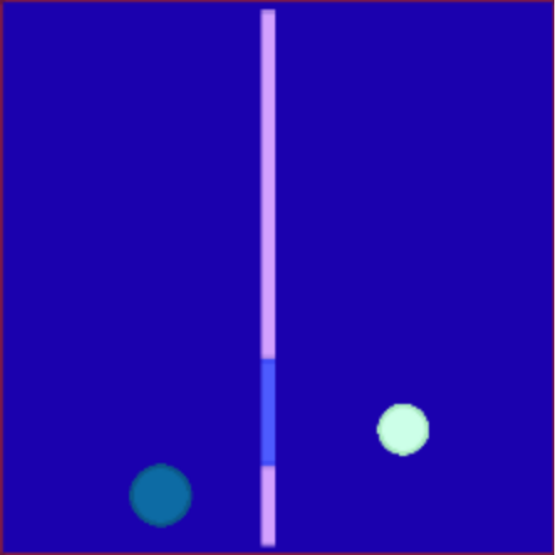}
      (b) TwoRoom
    \end{minipage}\hspace{0.3em}
    \begin{minipage}[t]{0.19\textwidth}
      \centering
     \includegraphics[width=\linewidth]{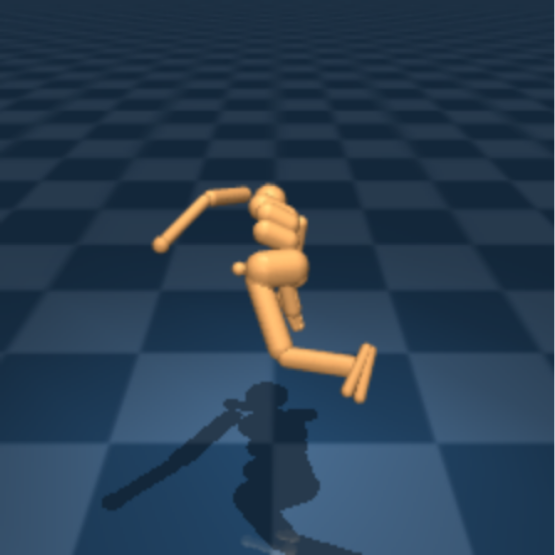}
     \includegraphics[width=\linewidth]{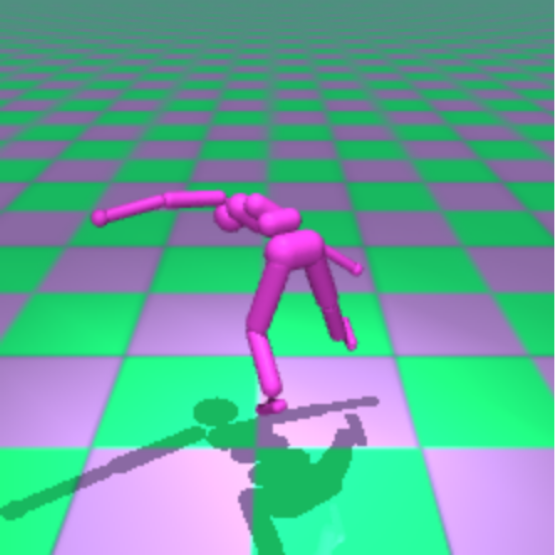}
      (c) DMC -- Humanoid
    \end{minipage}\hspace{0.3em}
    \begin{minipage}[t]{0.19\textwidth}
      \centering
     \includegraphics[width=\linewidth]{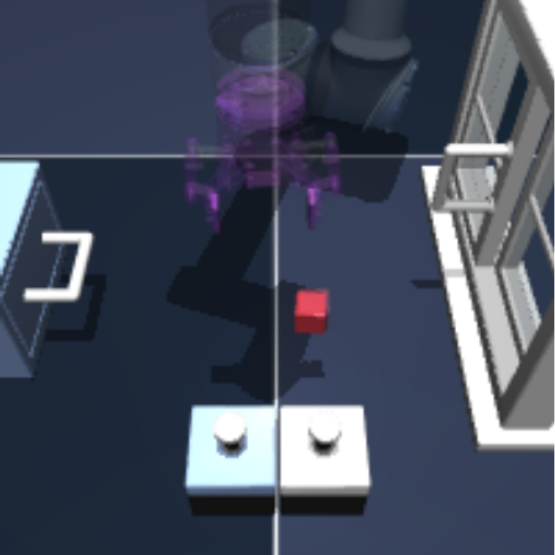}
     \includegraphics[width=\linewidth]{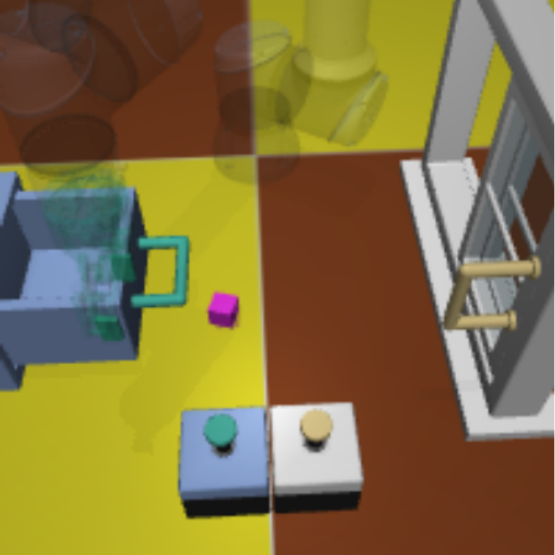}
      (d) OGBench -- Scene
    \end{minipage}
    \caption{\small\textbf{{\rm SWM} Environment Suite.} We support (and extend) a diverse set of established environments, including 2D/3D settings with tasks in manipulation, navigation, and classic control. (a) Push-T~\citep{chi2025diffusion}. A manipulation task where a blue agent needs to push a T-shaped block to match the green anchor. (b) Two-Room~\citep{sobal2025stresstesting}. A 2d navigation task where a red agent needs to navigate through a door to reach a green goal in the room. (c) DeepMind Control Suite~\citep{tassa2018deepmind}, a collection of 3d control tasks in MuJoCo. (d) OGBench~\citep{park2025ogbench}, a 3D robotic manipulation task collection in MuJoCo. (Top) Default settings. (Bottom) All factors of variations changing visual, geometric, and physical properties. All supported environments and their associated FoV can be found in Figure \ref{fig:all-env-appendix} and Table \ref{table:all-fov}.} 
    \label{fig:env-overview}
\end{figure}
{\rm SWM} is designed as a collection of diverse environments that span a wide range of design choices, including continuous and discrete state/action spaces, different action modalities, and varied agent embodiments. These environments differ not only in their task structure but also in their underlying dynamics or observation spaces, as illustrated in Figure \ref{fig:env-overview}. Such diversity allows evaluation across qualitatively distinct settings and supports broad comparisons of learning algorithms. However, evaluating generalization solely across different environments can obscure more fine-grained sources of variation that commonly arise within a single task or domain.

A key feature of {\rm SWM} is the notion of \textit{factors of variation (FoV)}. Each environment in the library exposes a set of optional controllable properties that enable systematic customization of the environment configuration. These factors of variation span multiple aspects, including visual attributes (e.g., color, shape, textures, lighting), geometric properties (e.g., size, orientation, position), and physical parameters (e.g., friction, damping, mass, gravity). By explicitly exposing these controls, {\rm SWM} enables fine-grained studies of robustness, generalization, domain shift, and continual learning within a single, unified environment. We provide a toy example in Listing \ref{lst:swm-fov}. More details about FoV can be found in Appendix \ref{appendix:swm-details}

\begin{lstlisting}[caption=\small{\textbf{{\rm SWM} Factor of Variation Logic.} During data collection or world reset, factors of variation (FoV) can optionally be specified via the \texttt{options} argument. In this illustrative Push-T example, all agent-related FoVs (e.g., color and size) are sampled, along with the color of the T-shaped object.}, label={lst:swm-fov}]
    import stable_worldmodel as swm
    
    world = swm.World('swm/PushT-v1', num_envs=2)
    world.set_policy(YourExpertPolicy())

    print(world.single_variation_space.names()) # available FoV

    # dataset with changing all agent FoV, and T color.
    world.record_dataset(
    dataset_name='pusht_demo',episodes=4, seed=0,
    options={"variation": ["agent", "block.color"]}, 
    )

\end{lstlisting}

Internally, FoVs are implemented as a new type of Gymnasium dictionary Space (in addition to the standard action and observation space), which stores an internal value that can be initialized, sampled with or without constraint.

\subsection{{\rm SWM} Evaluation Suite: Tasks, Planning Algorithms, and Baselines}

Evaluating world models is inherently challenging, as existing works rely on diverse evaluation settings. {\rm SWM} provides built-in support for \textit{goal-conditioned evaluation}, where the agent is tasked to reach a specified goal representation, such as a target state, image, or reward condition. Performance is measured in terms of success rate, defined as the percentage of evaluation episodes that end satisfying the goal condition.

In {\rm SWM}, evaluation can be conducted through the \texttt{World} interface and applied to the currently attached policy. These methods are \texttt{evaluate} and \texttt{evaluate\_from\_dataset}. {\rm SWM} is agnostic to the choice of policy. Yet, we provided some utilities to facilitate planning with Model Predictive Control (MPC)~\citep{richalet1978model} or Feed-Forward action prediction. We provided further details on the specifics of each evaluation method and different MPC solvers in Appendix \ref{appendix:swm-details}. 

\section{Experiments: DINO-WM Zero-Shot Robustness}
\label{sec:experiments}

\noindent
\begin{minipage}[t]{0.52\textwidth}
We now demonstrate how {\rm SWM} can be used as a research tool to analyze model robustness. Specifically, we leverage {\rm SWM} to evaluate the robustness of our reproduction of DINO-WM~\citep{zhou2025dino-wm} under both in-distribution and out-of-distribution evaluation settings, as well as its zero-shot generalization to environmental variations (e.g., agent color and background) in the Push-T environment. First, we observe that although DINO-WM performs well when evaluated on expert demonstrations, achieving a success rate of 94.0\%, its performance deteriorates sharply under distribution shift. When evaluated on reaching states drawn from trajectories collected by a random policy, the success rate drops to 12.0\%, revealing a strong dependence on the provenance of evaluation data. Next, using {\rm SWM} as a controlled evaluation framework, we probe DINO-WM’s zero-shot robustness to a range of factors of variation, as summarized in Table~\ref{tab:dino_wm_robustness}. Across all tested perturbations, the model exhibits consistently low scores, indicating limited robustness to unseen environmental variations despite the task structure remaining unchanged.

\end{minipage}
\hfill
\begin{minipage}[t]{0.45\textwidth}
\centering
\captionsetup{type=table}

\caption{\small\textbf{DINO-WM robustness on Push-T.} Zero-shot success rate on unseen FoVs, showing strong sensitivity to environment shifts.}
\label{tab:dino_wm_robustness}

\setlength{\tabcolsep}{5pt}
\label{tab:factor_variation_long}
\begin{tabular}{@{}l l c@{}}
\toprule
\textbf{FoV} & \textbf{Property} & \textbf{SR \% ($\uparrow$)} \\
\midrule
Color    & Anchor      & 20.0 \\
         & Agent       & 18.0 \\
         & Block       & 18.0 \\
         & Background  & 10.0 \\
\addlinespace
Size     & Anchor      & 14.0 \\
         & Agent       & 4.0  \\
         & Block       & 16.0 \\
\addlinespace
Angle    & Anchor      & 12.0 \\
         & Agent       & 12.0 \\
\addlinespace
Position & Anchor      & 4.0  \\
\addlinespace
Shape    & Agent       & 18.0 \\
         & Block       & 8.0  \\
\addlinespace
Velocity & Agent       & 14.0 \\
\midrule
None &        & 94.0 \\
\bottomrule
\end{tabular}
\end{minipage}


\section{Conclusion and Future Directions}
\label{sec:conclusion}

With a streamlined API {\rm SWM} promotes standardized evaluation, which we hope will accelerate progress in world-model research. We plan some future updates focusing on tools for improving debugging and interpretation of world models. 
Moreover, we will work on adding new environment support to the library with a focus on physical simulation or real-world tasks. Finally, our long-term vision aims to provide a standardized benchmark to keep track of the state-of-the-art in controllable world models, e.g., via a Hugging Face Benchmark.

\bibliography{iclr2026_conference}
\bibliographystyle{iclr2026_conference}

\appendix
\clearpage

\section{Code Example}
\label{appendix:code}

\subsection{End-to-End Pipeline.}

\begin{lstlisting}[caption={stable-worldmodel pipeline example}, label={lst:swm-pusht}]
import stable_worldmodel as swm
from stable_worldmodel.data import HDF5Dataset
from stable_worldmodel.policy import WorldModelPolicy, PlanConfig
from stable_worldmodel.solver import CEMSolver

world = swm.World('swm/PushT-v1', num_envs=8)
world.set_policy(your_expert_policy)

#=== Record Dataset ===
world.record_dataset(
    dataset_name='pusht_demo',
    episodes=100,
    seed=0,
    options={"variation": ["all"]},
)

# ... train your world model with pusht_demo ...
world_model = ...  # your world-model implementing get_cost

#=== Evaluate World Model ===
dataset = HDF5Dataset(
    name='pusht_demo',
    frameskip=1,
    num_steps=16,
    keys_to_load=['pixels', 'action', 'state']
)

# model predictive control
solver = CEMSolver(model=world_model, num_samples=300, device='cuda')
policy = WorldModelPolicy(
    solver=solver,
    config=PlanConfig(horizon=10, receding_horizon=5)
)

world.set_policy(policy)
results = world.evaluate(episodes=50, seed=0)

print(f"Success Rate: {results['success_rate']:.1f}%")
\end{lstlisting}

\subsection{Policy}

\begin{lstlisting}[caption={Policy definition and usage.}, label={lst:swm-policy}]
import stable_worldmodel as swm
class MyPolicy:
    def get_action(self, info: dict) -> np.ndarray:
        """
        Args:
            info: dict with all information collected from the environments
        Returns:
            actions: Array of shape (num_envs, action_dim)
        """
        return actions

# Use policy
world = swm.World('swm/PushT-v1', num_envs=8)
world.set_policy(MyPolicy())
\end{lstlisting}

\subsection{Dataset Recording}
\begin{lstlisting}[caption={{\rm SWM} Data collection.}, label={lst:swm-dataset}]
    import stable_worldmodel as swm
    
    world = swm.World('swm/PushT-v1', num_envs=8)
    world.set_policy(YourExpertPolicy())
    
    world.record_dataset(
    dataset_name='pusht_demo',
    episodes=100,
    seed=0,
    options={"variation": ["all"]},
    )

\end{lstlisting}

\section{{\rm SWM} Details}
\label{appendix:swm-details}
\subsection{Policy}

\paragraph{Policy.} Unlike Gymnasium, the \texttt{step} function does not take actions as an argument. Instead, actions are determined by a policy object associated with the world. At each call of the \texttt{step} method, the world queries the policy to obtain the actions for all environments. A policy is a simple Python object implementing a \texttt{get\_action} method. This method receives the current world \texttt{infos} and returns an action for each environment. Decoupling action selection from the \texttt{step} call makes it easy to swap policies within a single script without modifying the world interface. We provide a boilerplate example for policy implementation and usage in Listing \ref{lst:swm-policy}.

\paragraph{Model Predictive Control.} {\rm SWM} supports planning-based control by enabling world models to infer policies through the solution of a finite-horizon planning problem, i.e., optimizing the optimal sequence of actions reaching the goal. To this end, we provide a dedicated \texttt{MPCPolicy}. This policy is parameterized by a \texttt{PlanConfig}, which defines the Model Predictive Control (MPC) setup (e.g., planning horizon and receding horizon, warm start), and a \texttt{Solver} object responsible for optimizing the action sequence.

We re-implement several widely used planning solvers, including the Cross-Entropy Method (CEM), Model Predictive Path Integral (MPPI), and gradient-based optimizers (e.g., SGD, Adam). All solvers are implemented with efficiency and numerical stability in mind and are extensively tested to ensure reliability.

\subsection{Dataset Recording}

Once a world is created and a policy is attached, datasets can be collected using the \texttt{record\_dataset} method. This API runs episodes by executing the policy associated with the world and records the resulting interactions and all information contained in the internal state of the world. As a result, the quality and characteristics of the collected data are entirely determined by the chosen policy and world configuration. By default, we save all datasets in the HDF5 format. Yet, we support other formats like image folders or mp4 videos for specific usage. An illustrative example of dataset recording is provided in Listing \ref{lst:swm-fov}.

\subsection{Factor Of Variations}
FoVs are configured through an optional dictionary passed via the \texttt{options} argument at reset, dataset recording, or evaluation time. To enable variation, the \texttt{variations} key specifies a list of FoV names to be modified. We adopt a common hierarchical naming convention of the form \texttt{key\_1.key\_2} to reference FoVs within an environment. For example, \texttt{agent} applies variations to all agent-related properties, whereas \texttt{agent.color} restricts variation to the agent’s color only. All FoV can be changed simultaneously by setting \texttt{variations} to \texttt{all} as illustrated in listing \ref{lst:swm-dataset}. By default, specified FoVs are resampled at each \texttt{reset}; however, fixed values can be enforced by providing explicit assignments through the \texttt{variation\_values} key in \texttt{options}.

\subsection{Evaluations}

In {\rm SWM}, evaluation can be conducted under two complementary protocols, both accessible directly through the \texttt{World} interface and applied to the currently attached policy. 

First, an \textit{online} evaluation protocol samples (or allows the user to specify) both the initial state and the goal at the beginning of each episode, following prior work such as PLDM. This setting evaluates the policy through direct environment interaction and can be invoked using the world \texttt{evaluate} method.

Alternatively, {\rm SWM} supports an \textit{offline} evaluation protocol. In this setting, a complete trajectory is first sampled from a specified dataset, typically collected using an expert policy. The initial state and goal are then selected from this trajectory subject to a constraint on the maximum number of steps separating them. This protocol guarantees that the task is feasible within a given step budget, enabling controlled and reliable evaluation of planning and model accuracy without additional environment interaction. This setting, similar to DINO-WM, can be invoked using the world \texttt{evaluate\_from\_dataset} method.

\section{Experiment Details}
\label{appendix:exp-details}

\paragraph{Training Details.} Our re-implementation of DINO-WM has been implemented in PyTorch \citep{paszke2019pytorch} and trained with stable-pretraining \citep{balestriero2025stable}. We train for 20 epochs with the same hyperparameters as those prescribed in the original publication.

\paragraph{Evaluation Details.} We use the Cross Entropy Method (CEM) solver with the same set of parameters as the original DINO-WM publication. However, unlike the original work, which had an infinite planning budget, we fixed the steps budget to 50, which corresponds to 2x the minimum number of steps required to succeed (25).

\section{{\rm SWM} Environments}
\label{appendix:visu}

\begin{figure}
    \centering
    
    \begin{minipage}[t]{0.15\textwidth}
      \centering
     \includegraphics[width=\linewidth]{fig/swm_PushT.pdf}
     \includegraphics[width=\linewidth]{fig/swm_PushT_var.pdf}
      (a) PushT
    \end{minipage}\hspace{0.3em}
    \begin{minipage}[t]{0.15\textwidth}
      \centering
     \includegraphics[width=\linewidth]{fig/swm_TwoRoom.pdf}
     \includegraphics[width=\linewidth]{fig/swm_TwoRoom_var.pdf}
      (b) TwoRoom
    \end{minipage}\hspace{0.3em}
    \begin{minipage}[t]{0.15\textwidth}
      \centering
     \includegraphics[width=\linewidth]{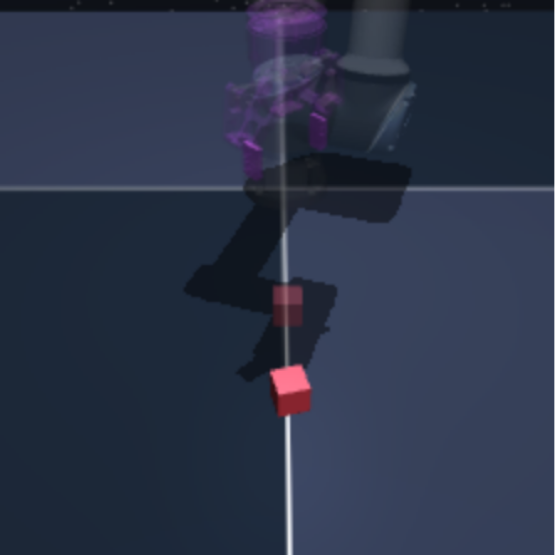}
     \includegraphics[width=\linewidth]{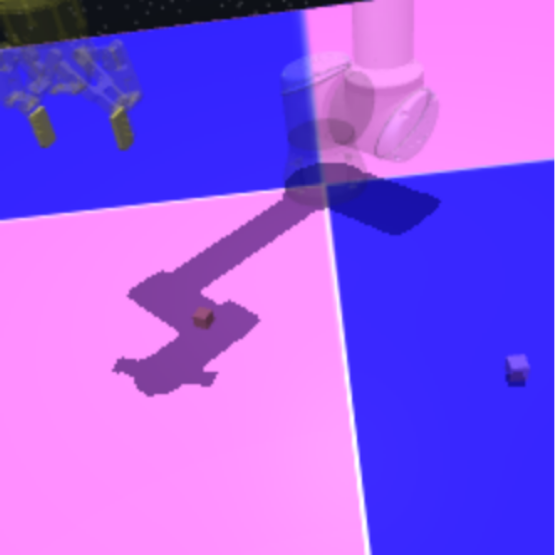}
      (c) OGBench -- Cube
    \end{minipage}\hspace{0.3em}
    \begin{minipage}[t]{0.15\textwidth}
      \centering
     \includegraphics[width=\linewidth]{fig/swm_OGBScene.pdf}
     \includegraphics[width=\linewidth]{fig/swm_OGBScene_var.pdf}
      (d) OGBench -- Scene
    \end{minipage}
    
    \vspace{0.6em}
    
    \begin{minipage}[t]{0.15\textwidth}
      \centering
     \includegraphics[width=\linewidth]{fig/swm_Humanoid.pdf}
     \includegraphics[width=\linewidth]{fig/swm_Humanoid_var.pdf}
      (e) Humanoid
    \end{minipage}\hspace{0.3em}
    \begin{minipage}[t]{0.15\textwidth}
      \centering
     \includegraphics[width=\linewidth]{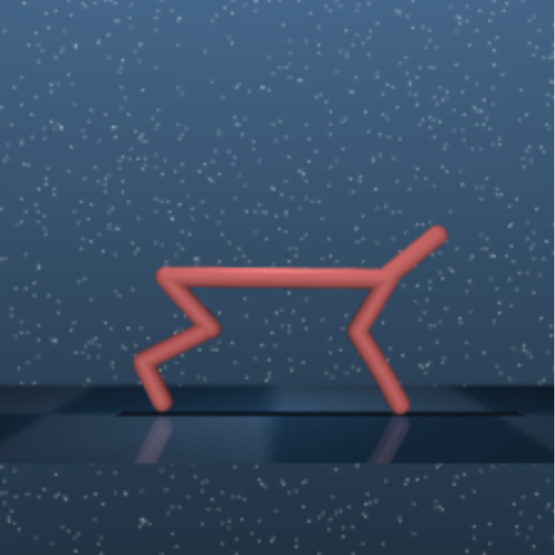}
     \includegraphics[width=\linewidth]{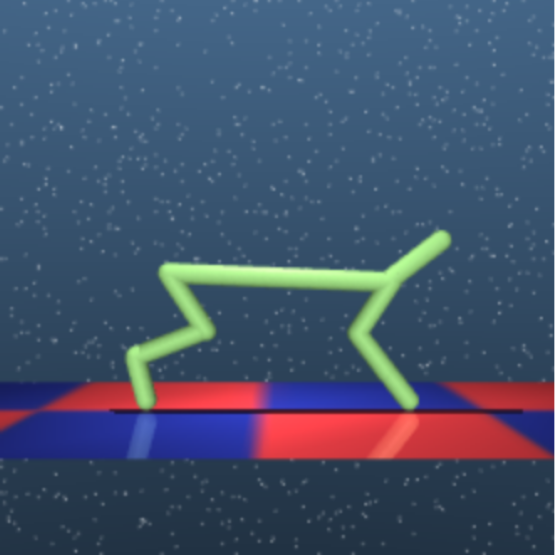}
      (f) Cheetah
    \end{minipage}\hspace{0.3em}
    \begin{minipage}[t]{0.15\textwidth}
      \centering
     \includegraphics[width=\linewidth]{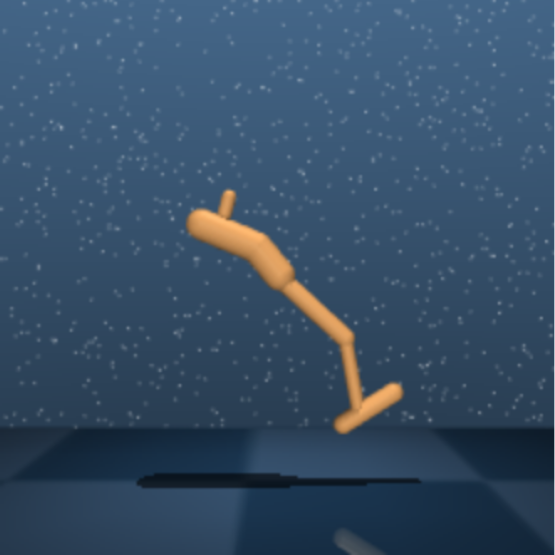}
     \includegraphics[width=\linewidth]{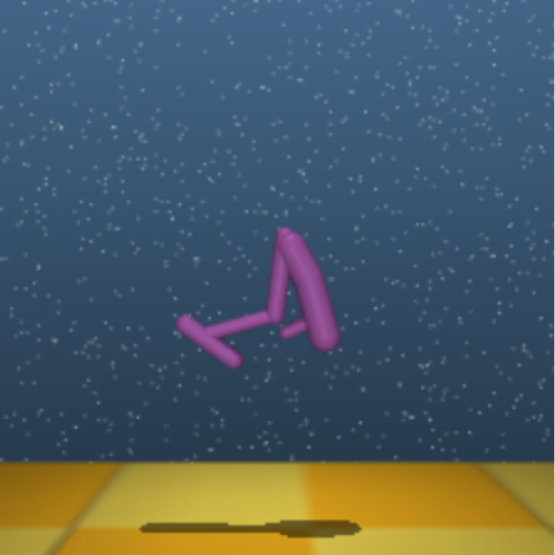}
      (g) Hopper
    \end{minipage}\hspace{0.3em}
    \begin{minipage}[t]{0.15\textwidth}
      \centering
     \includegraphics[width=\linewidth]{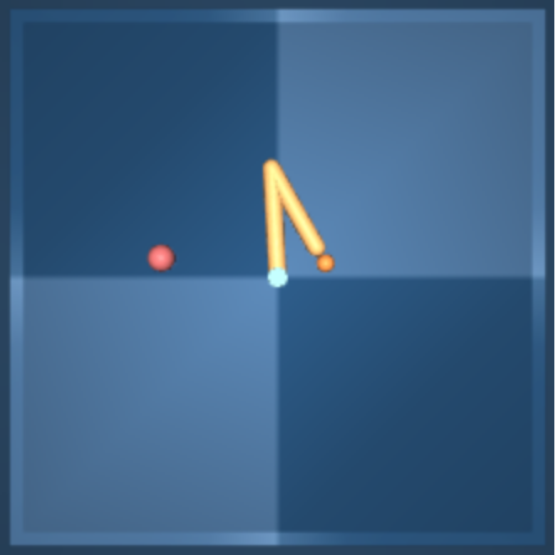}
     \includegraphics[width=\linewidth]{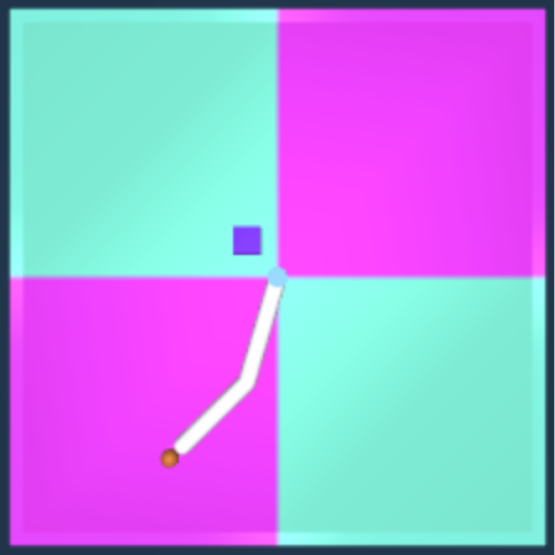}
      (h) Reacher
    \end{minipage}
    
    \vspace{0.6em}
    
    \begin{minipage}[t]{0.15\textwidth}
      \centering
     \includegraphics[width=\linewidth]{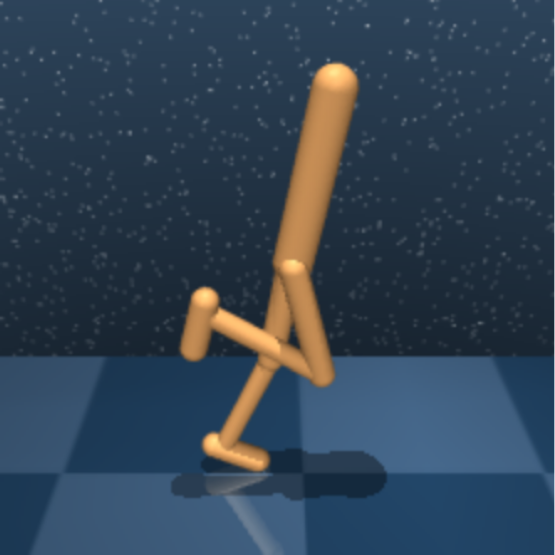}
     \includegraphics[width=\linewidth]{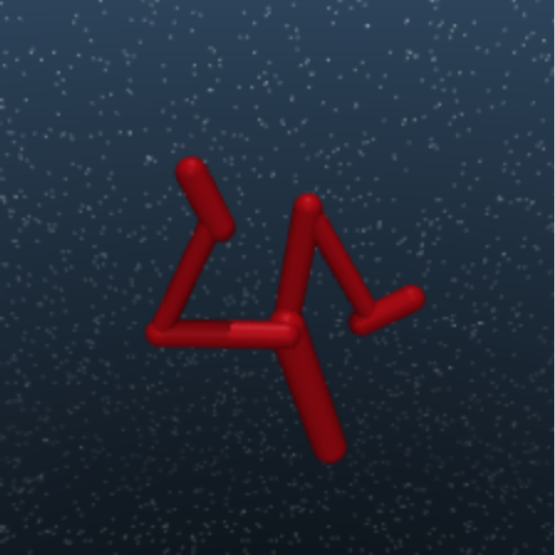}
      (i) Walker
    \end{minipage}\hspace{0.3em}
    \begin{minipage}[t]{0.15\textwidth}
      \centering
     \includegraphics[width=\linewidth]{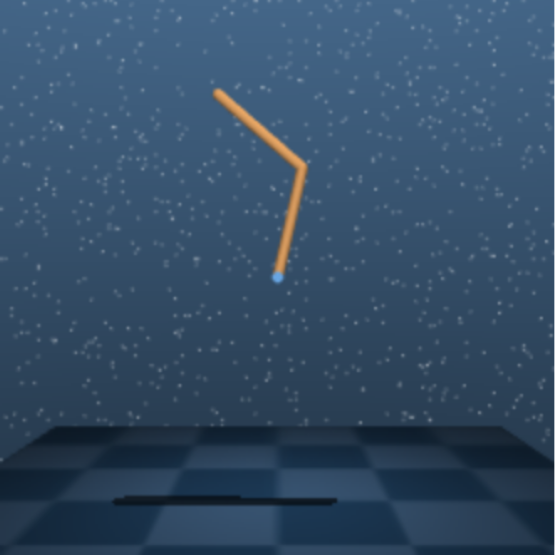}
     \includegraphics[width=\linewidth]{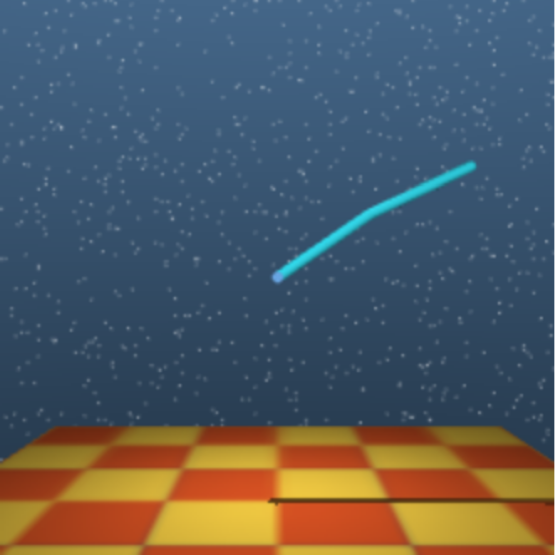}
      (j) Acrobot
    \end{minipage}\hspace{0.3em}
    \begin{minipage}[t]{0.15\textwidth}
      \centering
     \includegraphics[width=\linewidth]{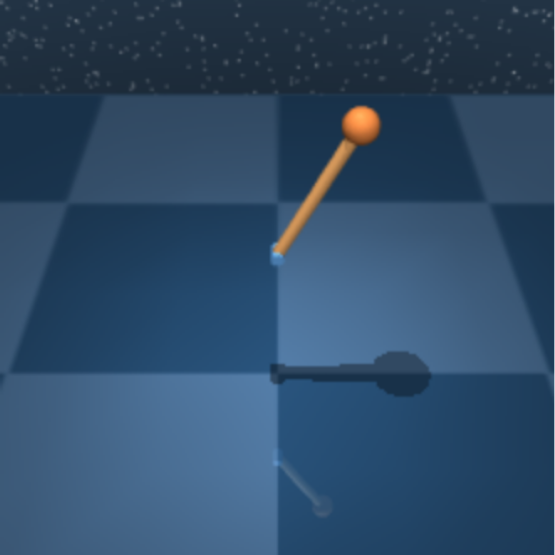}
     \includegraphics[width=\linewidth]{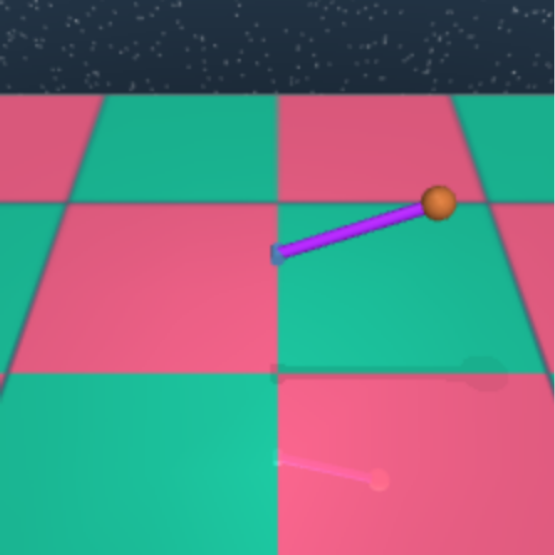}
      (k) Pendulum
    \end{minipage}\hspace{0.3em}
    \begin{minipage}[t]{0.15\textwidth}
      \centering
     \includegraphics[width=\linewidth]{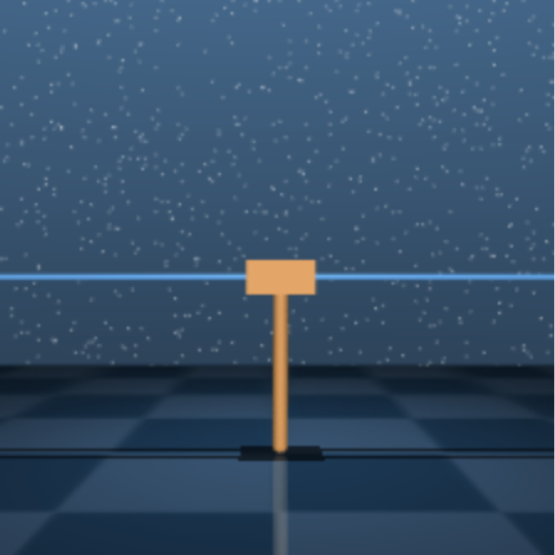}
     \includegraphics[width=\linewidth]{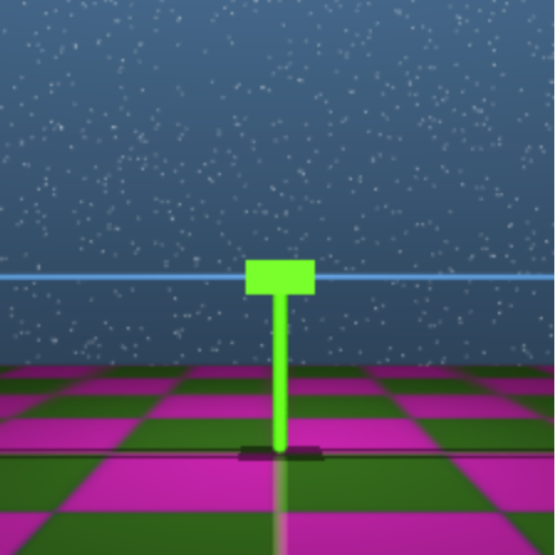}
      (l) Cartpole
    \end{minipage}
    
    \vspace{0.6em}
    
    \begin{minipage}[t]{0.15\textwidth}
      \centering
     \includegraphics[width=\linewidth]{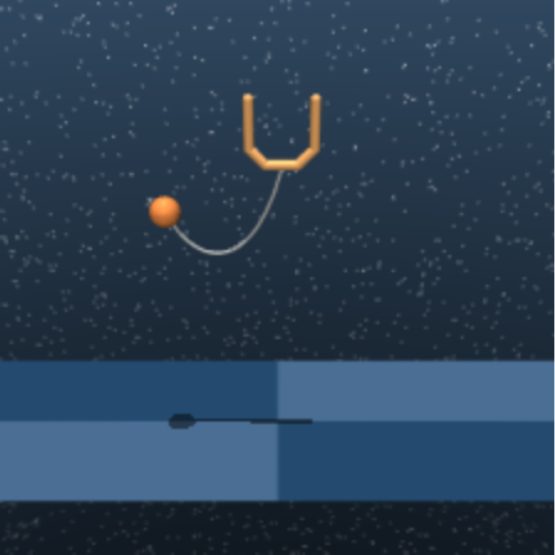}
     \includegraphics[width=\linewidth]{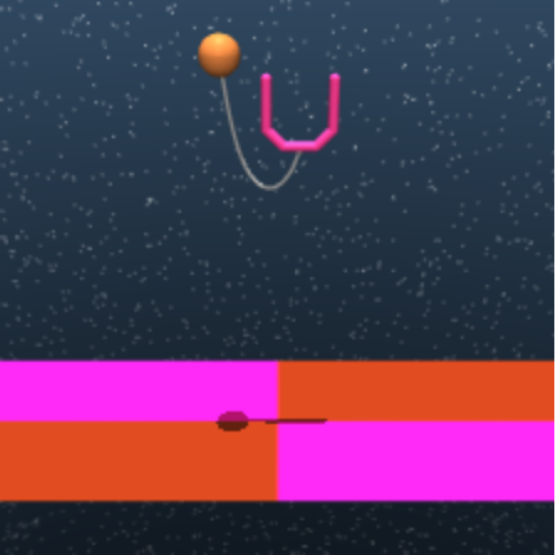}
      (m) Ball-in-Cup
    \end{minipage}\hspace{0.3em}
    \begin{minipage}[t]{0.15\textwidth}
      \centering
     \includegraphics[width=\linewidth]{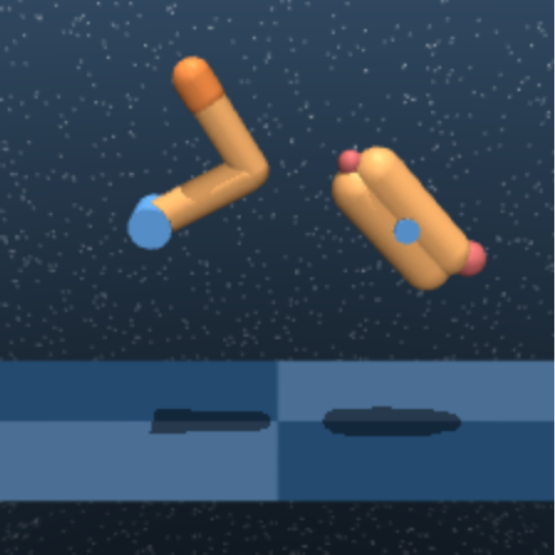}
     \includegraphics[width=\linewidth]{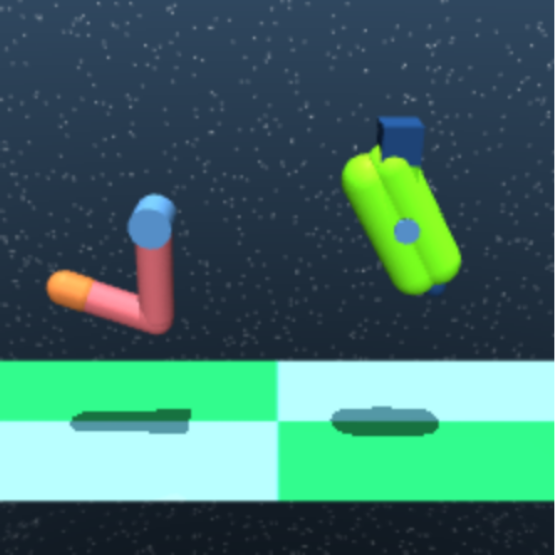}
      (n) Finger
    \end{minipage}\hspace{0.3em}
    \begin{minipage}[t]{0.15\textwidth}
      \centering
     \includegraphics[width=\linewidth]{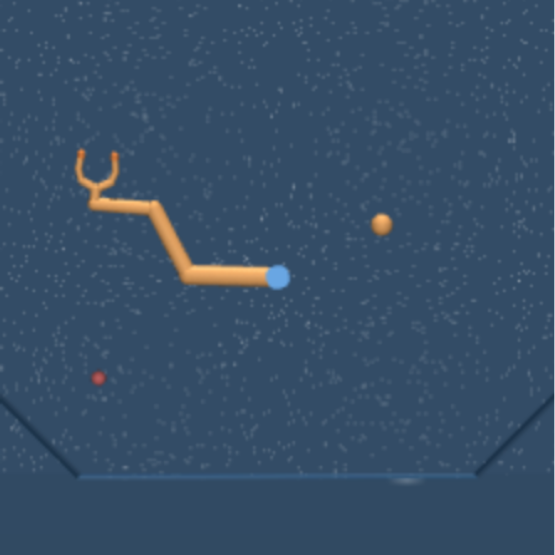}
     \includegraphics[width=\linewidth]{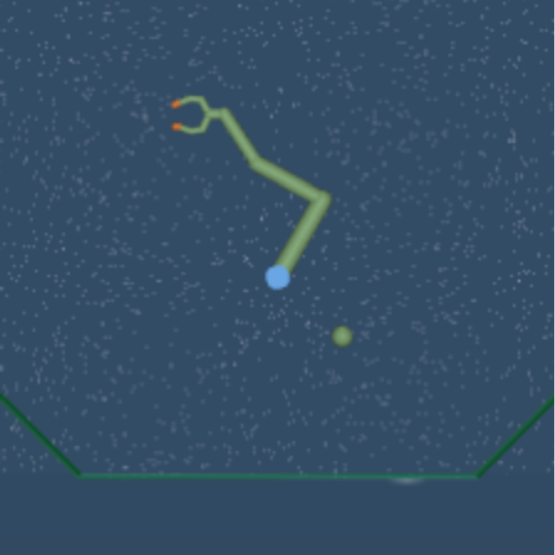}
      (o) Manipulator
    \end{minipage}\hspace{0.3em}
    \begin{minipage}[t]{0.15\textwidth}
      \centering
     \includegraphics[width=\linewidth]{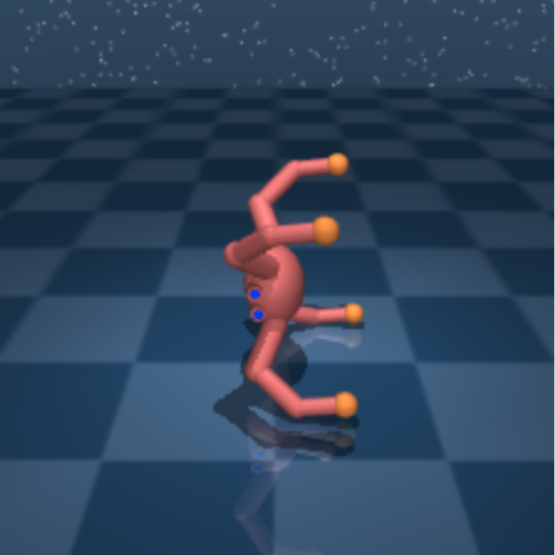}
     \includegraphics[width=\linewidth]{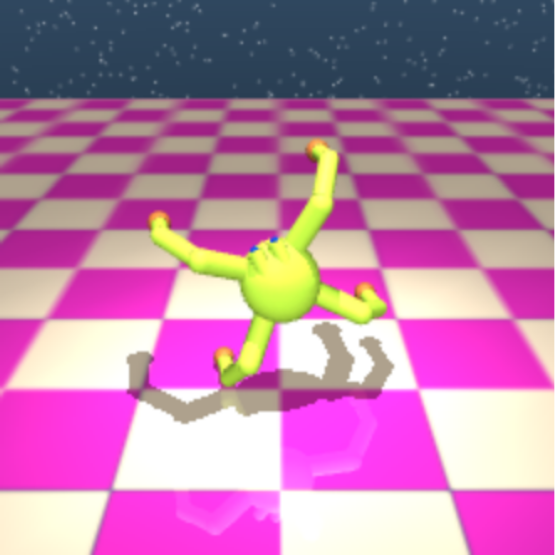}
      (p) Quadruped
    \end{minipage}

\caption{\small\textbf{Visualization of {\rm SWM} Environments suite.}}
\label{fig:all-env-appendix}
\end{figure}

\begin{table}[ht]
\centering
\small
\renewcommand{\arraystretch}{1.2}
\begin{tabular}{|l|c|p{9cm}|}
\hline
\textbf{Environment ID} & \textbf{\# FoV} & \textbf{ Available Variation(s)} \\
\hline
swm/PushT-v1 & 16 &
\texttt{agent.angle}, \texttt{agent.color}, \texttt{agent.scale}, \texttt{agent.shape},
\texttt{agent.start\_position}, \texttt{agent.velocity}, \texttt{background.color},
\texttt{block.angle}, \texttt{block.color}, \texttt{block.scale}, \texttt{block.shape},
\texttt{block.start\_position}, \texttt{goal.angle}, \texttt{goal.color},
\texttt{goal.position}, \texttt{goal.scale} \\
\hline
swm/TwoRoom-v1 & 17 &
\texttt{agent.color}, \texttt{agent.max\_energy}, \texttt{agent.position},
\texttt{agent.radius}, \texttt{agent.speed}, \texttt{background.color},
\texttt{door.color}, \texttt{door.number}, \texttt{door.position}, \texttt{door.size},
\texttt{goal.color}, \texttt{goal.position}, \texttt{goal.radius},
\texttt{wall.axis}, \texttt{wall.border\_color}, \texttt{wall.color},
\texttt{wall.thickness} \\
\hline
swm/OGBCube-v0 & 11 &
\texttt{agent.color}, \texttt{agent.ee\_start\_position}, \texttt{camera.angle\_delta},
\texttt{cube.color}, \texttt{cube.goal\_position}, \texttt{cube.goal\_yaw},
\texttt{cube.size}, \texttt{cube.start\_position}, \texttt{cube.start\_yaw},
\texttt{floor.color}, \texttt{light.intensity} \\
\hline
swm/OGBScene-v0 & 12 &
\texttt{agent.color}, \texttt{agent.ee\_start\_position}, \texttt{camera.angle\_delta},
\texttt{cube.color}, \texttt{cube.goal\_position}, \texttt{cube.goal\_yaw},
\texttt{cube.size}, \texttt{cube.start\_position}, \texttt{cube.start\_yaw},
\texttt{floor.color}, \texttt{light.intensity}, \texttt{lock\_color} \\
\hline
swm/HumanoidDMControl-v0 & 7 &
\texttt{agent.color}, \texttt{agent.left\_knee\_locked},
\texttt{agent.right\_lower\_arm\_density}, \texttt{agent.torso\_density},
\texttt{floor.color}, \texttt{floor.friction}, \texttt{light.intensity} \\
\hline
swm/CheetahDMControl-v0 & 7 &
\texttt{agent.back\_foot\_density}, \texttt{agent.back\_foot\_locked},
\texttt{agent.color}, \texttt{agent.torso\_density},
\texttt{floor.color}, \texttt{floor.friction}, \texttt{light.intensity} \\
\hline
swm/HopperDMControl-v0 & 7 &
\texttt{agent.color}, \texttt{agent.foot\_density}, \texttt{agent.foot\_locked},
\texttt{agent.torso\_density}, \texttt{floor.color},
\texttt{floor.friction}, \texttt{light.intensity} \\
\hline
swm/ReacherDMControl-v0 & 8 &
\texttt{agent.arm\_density}, \texttt{agent.color}, \texttt{agent.finger\_density},
\texttt{agent.finger\_locked}, \texttt{floor.color},
\texttt{light.intensity}, \texttt{target.color}, \texttt{target.shape} \\
\hline
swm/WalkerDMControl-v0 & 8 &
\texttt{agent.color}, \texttt{agent.left\_foot\_density},
\texttt{agent.right\_knee\_locked}, \texttt{agent.torso\_density},
\texttt{floor.color}, \texttt{floor.friction},
\texttt{floor.rotation\_y}, \texttt{light.intensity} \\
\hline
swm/AcrobotDMControl-v0 & 8 &
\texttt{agent.color}, \texttt{agent.lower\_arm\_density},
\texttt{agent.upper\_arm\_density}, \texttt{agent.upper\_arm\_locked},
\texttt{floor.color}, \texttt{light.intensity},
\texttt{target.color}, \texttt{target.shape} \\
\hline
swm/PendulumDMControl-v0 & 6 &
\texttt{agent.color}, \texttt{agent.mass\_density}, \texttt{agent.mass\_shape},
\texttt{agent.pole\_density}, \texttt{floor.color}, \texttt{light.intensity} \\
\hline
swm/CartpoleDMControl-v0 & 6 &
\texttt{agent.cart\_mass}, \texttt{agent.cart\_shape}, \texttt{agent.color},
\texttt{agent.pole\_density}, \texttt{floor.color}, \texttt{light.intensity} \\
\hline
swm/BallInCupDMControl-v0 & 9 &
\texttt{agent.color}, \texttt{agent.density}, \texttt{ball.color},
\texttt{ball.density}, \texttt{ball.size}, \texttt{floor.color},
\texttt{light.intensity}, \texttt{target.color}, \texttt{target.shape} \\
\hline
swm/FingerDMControl-v0 & 10 &
\texttt{agent.color}, \texttt{agent.fingertip\_density},
\texttt{agent.proximal\_density}, \texttt{floor.color},
\texttt{light.intensity}, \texttt{spinner.color}, \texttt{spinner.density},
\texttt{spinner.friction}, \texttt{target.color}, \texttt{target.shape} \\
\hline
swm/ManipulatorDMControl-v0 & 8 &
\texttt{agent.color}, \texttt{agent.hand\_density},
\texttt{agent.upper\_arm\_density}, \texttt{agent.upper\_arm\_length},
\texttt{floor.color}, \texttt{light.intensity},
\texttt{target.color}, \texttt{target.shape} \\
\hline
swm/QuadrupedDMControl-v0 & 7 &
\texttt{agent.color}, \texttt{agent.foot\_back\_left\_density},
\texttt{agent.knee\_back\_left\_locked}, \texttt{agent.torso\_density},
\texttt{floor.color}, \texttt{floor.friction}, \texttt{light.intensity} \\
\hline
\end{tabular}
\caption{Summary of {\rm SWM} environments, and controllable factor of variations.}
\label{table:all-fov}
\end{table}

\end{document}